\documentclass[conference]{IEEEtran}
\IEEEoverridecommandlockouts
\usepackage{cite}
\usepackage{amsmath,amssymb,amsfonts}
\usepackage{tablefootnote}
\usepackage{threeparttable}
\usepackage{algorithmic}
\usepackage{graphicx}
\def\BibTeX{{\rm B\kern-.05em{\sc i\kern-.025em b}\kern-.08em
    T\kern-.1667em\lower.7ex\hbox{E}\kern-.125emX}}
\usepackage{textcomp}
\usepackage{multirow}
\usepackage{tablefootnote}
\usepackage{xcolor}
\usepackage{hyperref}
\usepackage{etoolbox}
\usepackage{amsmath, amssymb}
\usepackage[inline]{enumitem}
\def\BibTeX{{\rm B\kern-.05em{\sc i\kern-.025em b}\kern-.08em
    T\kern-.1667em\lower.7ex\hbox{E}\kern-.125emX}}
\usepackage{orcidlink}

\usepackage{lipsum}
\usepackage{fancyhdr}
\fancypagestyle{sec}{\lhead{\tmpx}}

\fancypagestyle{arXiv}
{
   \fancyhf{}
   \chead{\textcolor{gray}{This paper will be presented as a lecture at the 2024 IEEE International Symposium on Circuits and Systems (ISCAS), Singapore, in the Special Session: RFIC \& AI: Pioneering New Wireless Communications}}
   \cfoot{ \fontsize{=9pt}{10pt}\selectfont  \textcolor{gray}{© 2024 IEEE. Personal use of this material is permitted. Permission from IEEE must be obtained for all other uses, in any current or future media, including reprinting/republishing this material for advertising or promotional purposes, creating new collective works, for resale or redistribution to servers or lists, or reuse of any copyrighted component of this work in other works}\\\fontsize{=10pt}{12pt}\selectfont\thepage}
   
}
\makeatletter
\patchcmd{\@maketitle}
  {\addvspace{0.5\baselineskip}\egroup}
  {\addvspace{-1\baselineskip}\egroup}
  {}
  {}
\makeatother

\begin{document}

\title{OpenDPD: An Open-Source End-to-End Learning \& Benchmarking Framework for Wideband Power Amplifier Modeling and Digital Pre-Distortion
}

\author{Yizhuo~Wu\orcidlink{0009-0009-5087-7349},
Gagan~Deep~Singh\orcidlink{0000-0002-0358-2451},
Mohammadreza~Beikmirza\orcidlink{0000-0002-3886-0554},
Leo~C.~N.~de~Vreede\orcidlink{0000-0002-5834-5461},\\
Morteza~Alavi\orcidlink{0000-0001-9663-5630}~and~Chang~Gao*\orcidlink{0000-0002-3284-4078}
\thanks{*Corresponding author: Chang Gao (chang.gao@tudelft.nl)} \\
\IEEEauthorblockA{Department of Microelectronics, Delft University of Technology, Delft, The Netherlands}
}

\maketitle

\begin{abstract}
With the rise in communication capacity, deep neural networks (DNN) for digital pre-distortion (DPD) to correct non-linearity in wideband power amplifiers (PAs) have become prominent. Yet, there is a void in open-source and measurement-setup-independent platforms for fast DPD exploration and objective DPD model comparison. This paper presents an open-source framework, \texttt{OpenDPD}, crafted in \texttt{PyTorch}, with an associated dataset for PA modeling and DPD learning. We introduce a Dense Gated Recurrent Unit (DGRU)-DPD, trained via a novel end-to-end learning architecture, outperforming previous DPD models on a digital PA (\textbf{DPA}) in the new digital transmitter (DTX) architecture with unconventional transfer characteristics compared to analog PAs. Measurements show our DGRU-DPD achieves an ACPR of -44.69/-44.47\,dBc and an EVM of -35.22\,dB for 200\,MHz OFDM signals. \texttt{OpenDPD} code, datasets and documentation are publicly available at \textcolor{red}{\url{https://github.com/lab-emi/OpenDPD}}.
\end{abstract}


\section{Introduction}
\thispagestyle{arXiv}


In the evolution of 6G and Wi-Fi 7 wireless systems and beyond, the need for enhanced Digital Pre-distortion (\textbf{DPD}) is underscored by growing communication capacities and data accuracy demands~\cite{wesemann2023energy}. DPD mitigates the non-linearity of Power Amplifiers (\textbf{PA}) by deriving its inverse function. While traditional methods like the general memory polynomial (\textbf{GMP})~\cite{GMP} are popular in the industry, the increasingly wider signal bandwidths introduce more intricate non-linearity and memory effects in the PA, presenting challenges to conventional linearization techniques.

Deep learning has catalyzed significant advances in DNN-based DPD for wideband PAs. The Time Delay Neural Network (\textbf{TDNN}) was the earliest DNN architecture explored to model PA's memory effects~\cite{RVTDNN,RVFTDNN,VDTDNN,ARVTDNN}. Gated RNNs, notably LSTM~\cite{Hochreiter1997LSTM} and GRU~\cite{Cho2014GRU}, excel in sequence modeling for state-of-the-art DPDs. Integrating vector decomposition (\textbf{VD}) with LSTM led to VDLSTM~\cite{VDLSTM}, while Phase-Gated JANET~\cite{PGJANET} enhanced linearization by combining an amplitude-phase extractor with JANET~\cite{Westhuizen2018JANET}. The DVR technique with JANET, explored in~\cite{DVRJANET}, has further improved linearization for signals up to 200\,MHz. Meanwhile, CNNs have been used to analyze PA signals as images to boost linearization~\cite{Hu2022RVTDCNN}. However, the improvements in DPD models entail greater computational demand and model expansion.

In many deep learning fields, standardized datasets direct research. For example, MNIST~\cite{LeCun2010MNIST} is fundamental for image classification, while Penn Treebank~\cite{Marcus1994Penn} is pivotal in language modeling. Yet, the DPD field presents inconsistent experimental setups and diverse learning methodologies. This variability complicates objective evaluations of DPD models and emphasizes the need for a benchmark similar to MNIST.

Regarding the discrepancy in DPD learning architectures, traditional methods present notable challenges. The Indirect Learning Architecture (\textbf{ILA}) hinges on the assumption of commutability of PAs~\cite{ILA}. Conversely, the Direct Learning Architecture (\textbf{DLA}) faces difficulty obtaining the ideal pre-distorted PA input~\cite{DLA}. Although the Iterative Learning Control (\textbf{ILC}) addresses the challenge of pinpointing ideal values within DLA, it introduces added complexity through the necessity of multiple measurement iterations~\cite{ILC}. Considering experimental setups, prior studies use varied PAs. Although RF-WebLab~\cite{Landin2015WebLab} is publicly accessible, it focuses on analog PA. The emerging digital transmitter (\textbf{DTX})~\cite{Alavi2014TMTT,Jung2020JSSC,Beikmirza2021JSSC,Mul2023TMTT} architecture remains underexplored despite its unique transfer characteristics~\cite{Beikmirza2023}.

This work presents the main contributions below:
\begin{enumerate}
    \item We present \texttt{OpenDPD}, an open-source, end-to-end (\textbf{E2E}) learning framework crafted in \texttt{PyTorch}~\cite{Paszke2019PyTorch}, streamlining the prototyping of non-linearity correction methods and establishes a standardized benchmark for DNN-based DPD models.
    
    \item The \texttt{OpenDPD} platform offers a public dataset with I/Q modulated signals from a digital PA (\textbf{DPA}) in a DTX. Additionally, \texttt{OpenDPD} features functions to evaluate simulated (\textbf{SIM})-ACPR and SIM-EVM using a provided pre-trained PA model. This not only simplifies DPD design exploration but also ensures consistent testing conditions, enabling researchers to make objective comparisons across DPD studies.

    \item We present the Dense GRU (\textbf{DGRU}) for PA modeling and DPD, surpassing the performance of LSTM, GRU, and prior works in DTX non-linearity correction regarding the measured Adjacent Channel Power Ratio (\textbf{ACPR}) and Error Vector Magnitude (\textbf{EVM}).
    
    \item \texttt{OpenDPD} is designed to be extensible, with plans to encompass datasets from diverse PAs, and we cordially invite contributions from various research disciplines.
\end{enumerate}
\begin{figure}[t]
        \centering
        \includegraphics[width=0.9\linewidth]{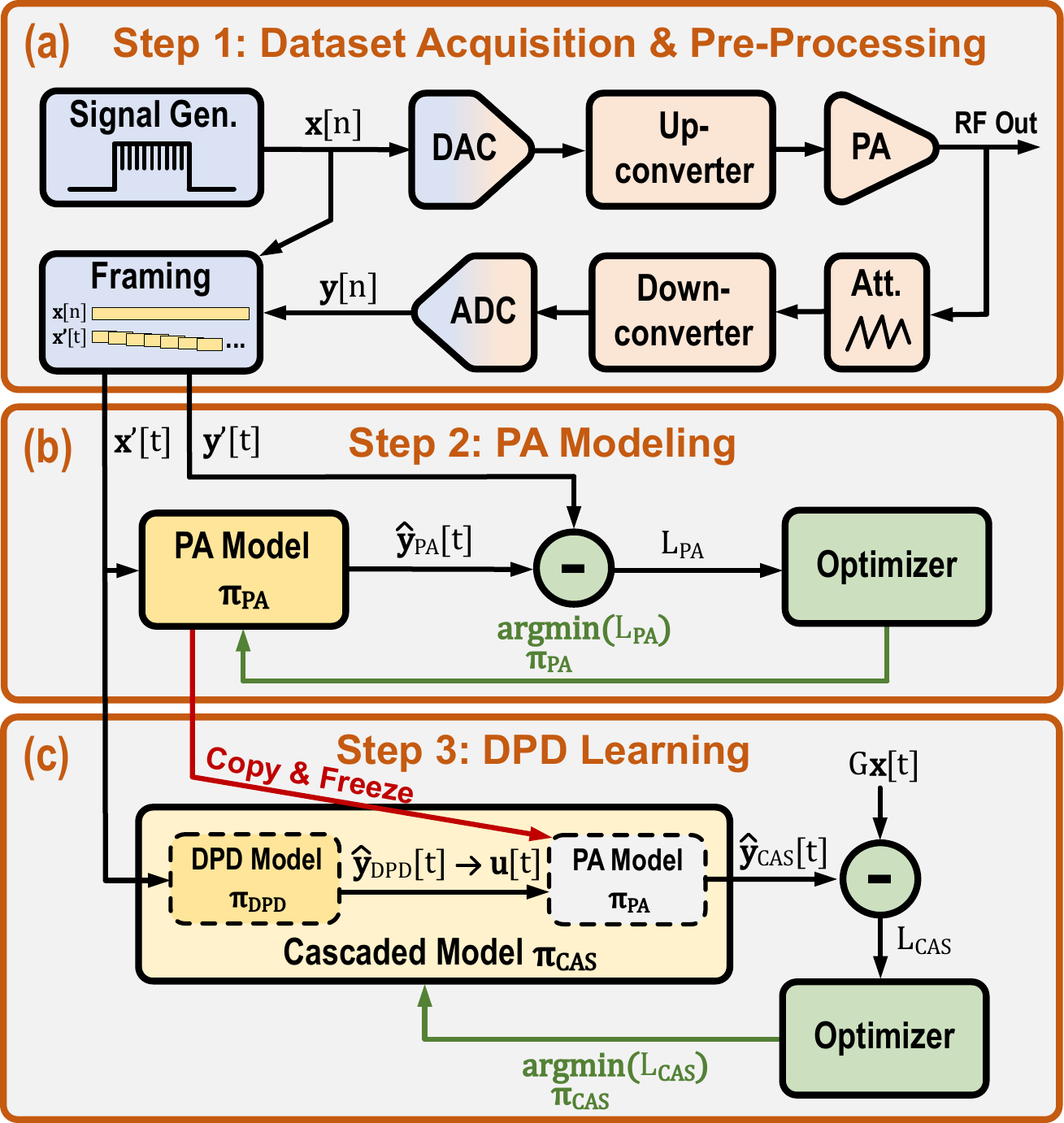}
        \caption{Generalized E2E learning architecture. With a DPA, an RF-DAC will be used instead of the DAC, Up-converter, and analog PA.}
        \label{fig:E2E}
    \end{figure}
\section{The \texttt{OpenDPD} Framework}
\label{sec:framework}
\subsection{E2E Learning}
We introduce the E2E learning architecture that uses backpropagation to train a DPD model with a pre-trained PA behavioral model, eliminating the assumption of a commutable PA in ILA and avoiding the manual determination of optimal pre-distorted PA inputs found in DLA. The idea of backpropagation through a DNN-based PA model is inspired by~\cite{E2E}. However, previous studies did not evaluate this on wideband PA nor report concrete measurements in ACPR and EVM.

As depicted in Fig.~\ref{fig:E2E}, the E2E learning architecture consists of three primary steps:
\begin{enumerate}
    \item \textbf{Data Acquisition \& Pre-Processing (Fig.~\ref{fig:E2E}a):} The baseband I/Q signal $\mathbf{X} = \{\mathbf{x}[n] | \mathbf{x}[n] = I_\mathbf{x}[n] + jQ_\mathbf{x}[n], I_\mathbf{x}[n], Q_\mathbf{x}[n] \in \mathbb{R}, n \in 0,\dots ,N-1\}$ is sent to the PA. The output signals (labels), $\mathbf{Y} = \{\mathbf{y}[n]| \mathbf{y}[n] = I_\mathbf{y}[n] + jQ_\mathbf{y}[n], I_\mathbf{y}[n], Q_\mathbf{y}[n] \in \mathbb{R}, n \in 0,\dots,N-1\}$, are sourced from the PA under test. 
    To address gradient vanishing issues and enhance training, feature and label sequences are split into shorter frames, $\mathbf{X}'$ and $\mathbf{Y}'$, each of size $T$, with a stride of $S$. Consecutive frames overlap by $T-S$ samples.
    
    \item \textbf{PA Modeling (Fig.~\ref{fig:E2E}b):} Utilizing framed inputs $\mathbf{x}'[t]$ and framed target outputs $\mathbf{y}'[t]$, a behavioral PA model, $\pi_{PA}$, is trained in a sequence-to-sequence learning way via BackPropagation Through Time (\textbf{BPTT}). The objective is to minimize the Mean Squared Error (\textbf{MSE}) loss function given by $L_{PA} = \frac{1}{B \times T} \sum_{b=1}^{B} \sum_{t=0}^{T-1} (\hat{\mathbf{y}}_{PA}[t] - \mathbf{y}'[t])^2$, where $B$ is the batch size in the mini-batched stochastic gradient descent and $\hat{\mathbf{y}}_{PA}[t]$ is the prediction output of the PA model.
    
    \item \textbf{DPD Learning (Fig.~\ref{fig:E2E}c):} Before the pre-trained PA behavioral model $\pi_{PA}$, a DPD model, $\pi_{DPD}$, is instantiated and its output $\hat{\mathbf{y}}_{DPD}[t]$ is fed to the input of $\pi_{PA}$ to form a cascaded model $\pi_{CAS}$. In this configuration, the parameters of $\pi_{PA}$ remain unaltered. The learning process involves feeding the framed inputs $\mathbf{x}'[t]$ through $\pi_{CAS}$ and executing BPTT across both $\pi_{PA}$ and $\pi_{DPD}$ to minimize $L_{CAS} = \frac{1}{B \times T} \sum_{b=1}^{B} \sum_{n=0}^{T-1} (\hat{\mathbf{y}}_{CAS}[t] - G\mathbf{x}[t])^2$, where $G$ is the target gain of the cascaded DPD-PA system.
\end{enumerate}

After the learning process, $\hat{\mathbf{y}}_{DPD}$ converges to the ideal PA inputs $\mathbf{U} = \{\mathbf{u}[n] | \mathbf{u}[n] = I_\mathbf{u}[n] + jQ_\mathbf{u}[n], I_\mathbf{u}[n], Q_\mathbf{u}[n] \in \mathbb{R}, n \in 0,\dots ,N-1\}$.
\begin{figure}[t]
    \centering
    \includegraphics[width=0.9\linewidth]{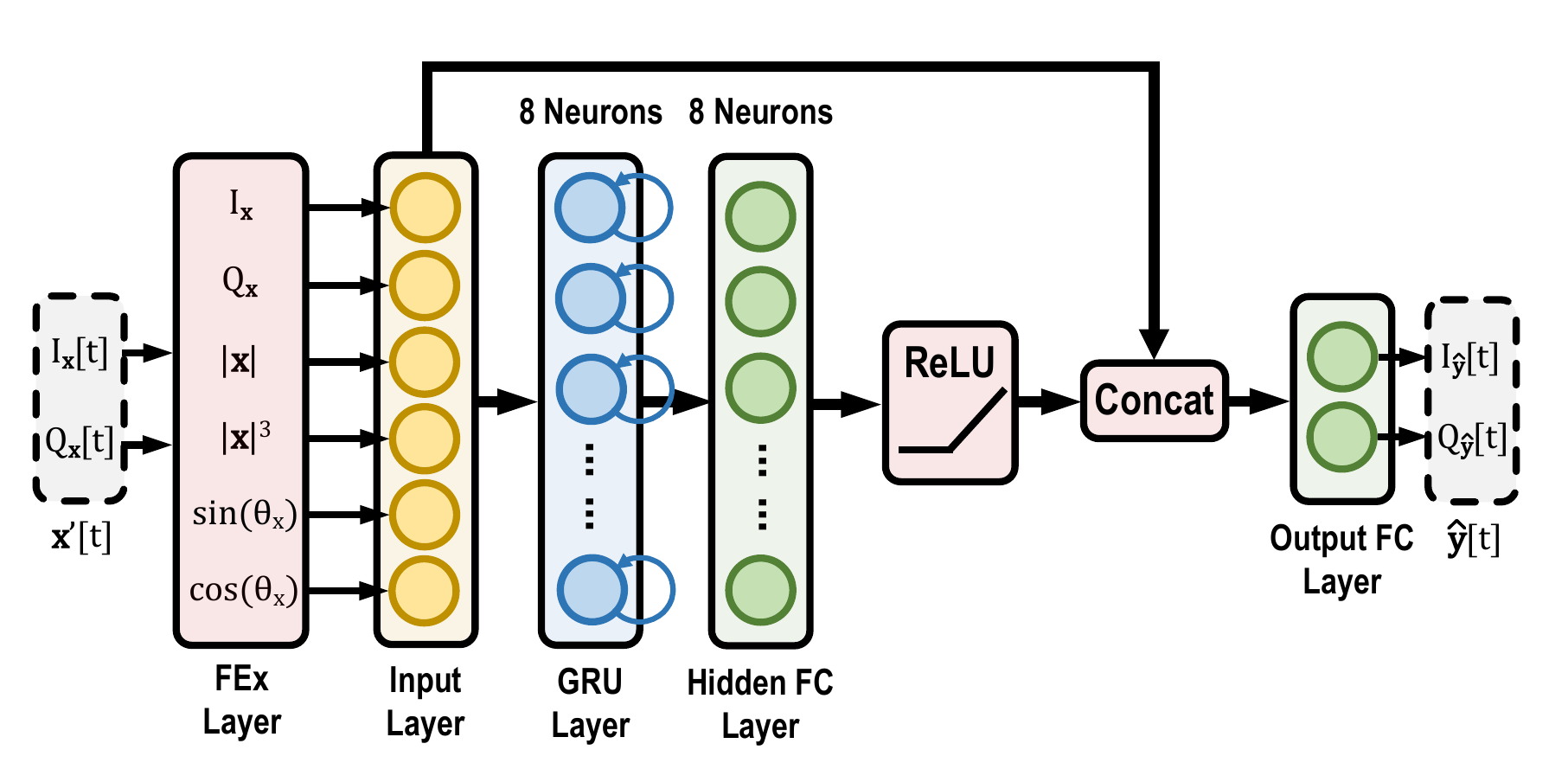}
    \caption{The DGRU-DPD architecture.}
    \label{fig:DGRU}
\end{figure}
\begin{figure}[t]
    \centering
    \includegraphics[width=0.65\linewidth]{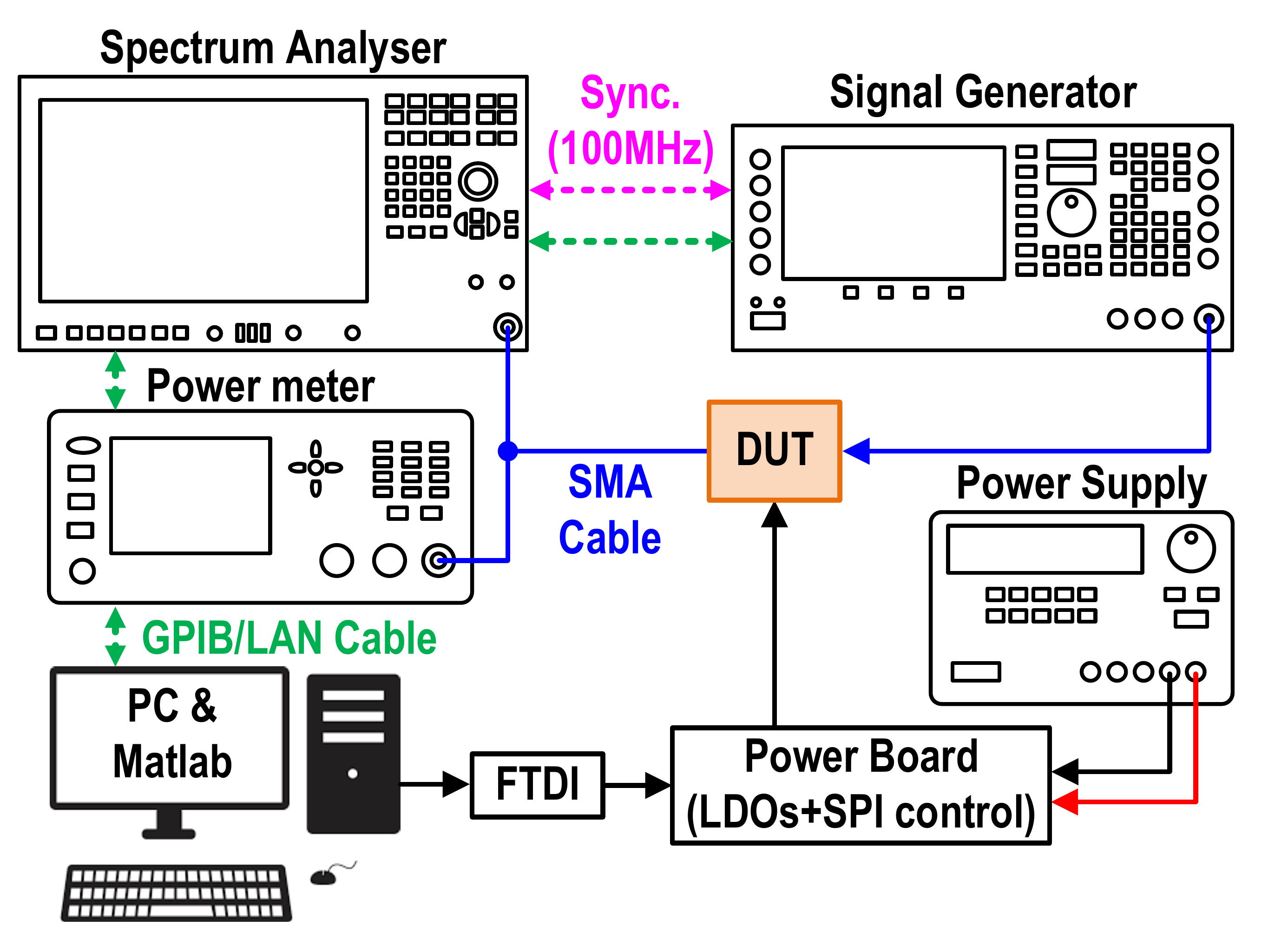}
    \caption{Setup for dataset acquisition and DPD performance measurement.}
    \label{fig:platform}
\end{figure}
\begin{figure*}[t]
    \centering
    \includegraphics[width=1.0\linewidth]{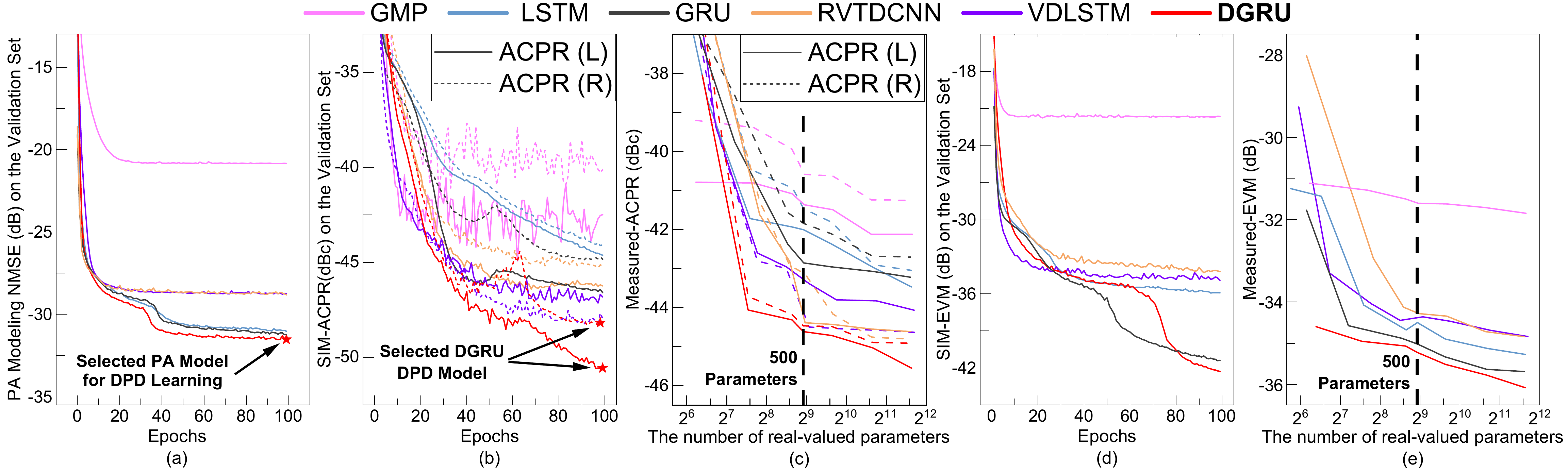}
    \caption{Training and measurement results on 200\,MHz 10-channel$\times$20\,MHz OFDM signals from the \texttt{DPA\_200MHz} validation set. Each curve represents the best performance of each algorithm over 5 random seeds. (a) The 500-parameter PA modeling NMSE over training epochs. The gold PA model of each algorithm is saved at the lowest NMSE; (b) The 500-parameter DPD learning SIM-ACPR over training epochs. The gold DPD model of each algorithm is saved at the lowest averaged SIM-ACPR; (c) Measured ACPR vs. real-valued model parameters; (d) The 500-parameter DPD learning SIM-EVM over training epochs; (e) Measured EVM vs. real-valued model parameters.}
    \label{fig:ACPR}
\end{figure*}
\subsection{Benchmarking}
\texttt{OpenDPD} has a standardized dataset and benchmarks, including SIMulated (\textbf{SIM})-ACPR and SIM-EVM, for fair comparison of DPD models across different machine learning approaches, enhancing quick DPD prototyping.

\subsubsection{Dataset}
The \texttt{DPA\_200MHz} dataset uses 200\,MHz 10-channel $\times$ 20\,MHz I/Q modulated Orthogonal Frequency Division Multiplexing (\textbf{OFDM}) signals, with each channel using 64-QAM and 64 subcarriers. It is partitioned into training (60\%), validation (20\%), and test (20\%) sets, containing the DPA's input-output signals, as captured in Fig.~\ref{fig:platform}. 

\subsubsection{Comparison between DPD Models}
\texttt{OpenDPD} users should train the PA behavioral and DPD models using the training set, optimizing hyperparameters like learning rate and batch size with the validation set, keeping test data unused. The platform calculates NMSE, SIM-ACPR, and SIM-EVM as follows:
\begin{align}
&\text{NMSE} = \frac{\sum_{n=0}^{N-1} \left( (I_{\hat{y}}[n] - I_{y}[n])^2 + (Q_{\hat{y}}[n] - Q_{y}[n])^2 \right)}{\sum_{n=0}^{N-1} (I_{y}[n]^2 + Q_{y}[n]^2)}, \\
&\text{ACPR and SIM-ACPR} =  \frac{P_{\text{adj}}}{P_{\text{main}}},
\end{align}
where $P_{main}$ and $P_{adj}$ are the main and adjacent channel powers, respectively. ACPR is measured while SIM-ACPR is software-derived. For EVM:
{\small
\begin{align}
    &\text{EVM} =  \frac{\sum_{k=0}^{K-1} \left( (I_{y,\text{mea}}[k] - I_{x}[k])^2 + (Q_{y,\text{mea}}[k] - Q_{x}[k])^2 \right)}{\sum_{k=0}^{K-1} \left( I_{x}[k]^2 + Q_{x}[k]^2 \right)},
\end{align}}
with $I[k]$ and $Q[k]$ being the FFT-derived in-phase and quadrature signal components and "mea" denoting measured PA outputs. SIM-EVM is analogously computed in software.
\subsection{Dense GRU-based DPD}
We introduce a novel DGRU-DPD architecture shown in Fig.~\ref{fig:DGRU}. This architecture has a dense skip path with the conventional GRU-RNN structure to concatenate features to the input of the output fully connected (\textbf{FC}) layer. The motivation is to mitigate the gradient vanishing issue and reuse input features to augment the linearization efficacy of the model~\cite{Huang2017DenseNet}. DGRU equips an online Feature Extractor (FEx) calculating $|x[t]|$, $|x[t]|^3$, $\text{sin}\left(\theta_\mathbf{x}[t]\right)=Q_\mathbf{x}[t]/|x[t]|$, and $\text{cos}\left(\theta_\mathbf{x}[t]\right)=I_\mathbf{x}[t]/|x[t]|$ from $I_\mathbf{x}[t]$, $Q_\mathbf{x}[t]$ and sending them to downstream layers. The \texttt{OpenDPD} platform offers a pre-trained DGRU model, facilitating ease of reproduction, deployment, and fine-tuning by subsequent research endeavors.

\section{Experimental Results}
\label{sec:exp}
\subsection{Experimental Setup, Dataset and Training}
The experimental setup is shown in Fig.~\ref{fig:platform}. The I/Q data was processed using a 40\,nm CMOS-based DTX~\cite{Beikmirza2023JSSC} and then digitized by an R\&S-FSW8 analyzer. The DPA's average power is $13.42$\,dBm, and the PAPR of the signal without DPD is about 9.6\,dB. With DPD, the PAPR of the signal is about 11\,dB. \texttt{OpenDPD} has the flexibility to set the target gain $G$ to arbitrary values. In this work, we particularly used:
\begin{equation}
    G = \frac{1}{N} \sum_{n=0}^{N-1}\frac{\sqrt{ I_y[n]^2 + Q_y[n]^2}}{\sqrt{I_x[n]^2 + Q_x[n]^2}}
\end{equation}
\begin{table*}[t]
\centering
\begin{threeparttable}
\caption{Performance Comparison of DPD Models with Approximately 500 Real-Valued Parameters on 200\,MHz 10-channel\(\times\)20\,MHz OFDM Signals from \texttt{DPA\_200MHz} Test Set Averaged on 5 Random Seeds $\pm$ Standard Deviations.}
\label{tab:200}
\begin{tabular}{|c|c|c|cc|cc|}
\hline
\textbf{Classes} & \multicolumn{1}{|c|}{\textbf{DPD Models}\tnote{a}} & \multicolumn{1}{c|}{\begin{tabular}[c]{@{}c@{}}\textbf{DPD-NMSE}\tnote{b}\\ \textbf{(dB)}\end{tabular}} & \multicolumn{1}{c}{\begin{tabular}[c]{@{}c@{}}\textbf{SIM-ACPR} \\ \textbf{(dBc, L/R)}\end{tabular}} & \multicolumn{1}{c|}{\begin{tabular}[c]{@{}c@{}}\textbf{Measured-ACPR}\\ \textbf{(dBc, L/R)}\end{tabular}} & \begin{tabular}[c]{@{}c@{}}\textbf{SIM-EVM}\\ \textbf{(dB)}\end{tabular} & \begin{tabular}[c]{@{}c@{}}\textbf{Measured-EVM}\\ \textbf{(dB)}\end{tabular} \\ \hline \hline
Without DPD & - & - & -32.19$\pm$0.67/-30.71$\pm$0.75\tnote{c} & -32.74/-31.71 & -33.06$\pm$0.73\tnote{c} & -27.36\\ \hline \hline
\multirow{3}{*}{Baseline Models} & GMP~\cite{GMP} & -21.07$\pm$0.06 & -42.60$\pm$1.40/-39.72$\pm$1.28 & -41.37$\pm$0.67/-40.59$\pm$0.24 & -21.50$\pm$0.27 & -31.60$\pm$0.30 \\  
 & LSTM & -36.27$\pm$2.73 & -44.39$\pm$0.78/-42.80$\pm$0.80 & -42.01$\pm$0.58/-41.47$\pm$0.86 & -35.98$\pm$2.79 & -34.50$\pm$0.81 \\ 
 & GRU & -39.63$\pm$0.51 & -44.84$\pm$1.03/-43.79$\pm$0.52 & -42.86$\pm$0.53/-41.85$\pm$1.01 & -42.66$\pm$1.10 & -35.02$\pm$0.62 \\ \hline \hline
\multirow{2}{*}{Previous Works} & RVTDCNN~\cite{Hu2022RVTDCNN} & -31.63$\pm$0.19 & -44.51$\pm$0.97/-44.39$\pm$0.29 & -44.39$\pm$0.47/-43.14$\pm$0.47 & -33.80$\pm$0.28 & -33.85$\pm$0.21 \\ 
 & VDLSTM~\cite{VDLSTM} & -32.15$\pm$0.15 & -45.42$\pm$0.79/-45.92$\pm$0.32 & -43.39$\pm$0.40/-44.50$\pm$0.71 & -34.27$\pm$0.12 & -34.42$\pm$0.16 \\ \hline \hline
\multirow{2}{*}{\textbf{This Work}} & Features\tnote{d}\,\,+\,GRU & -39.84$\pm$2.23 & -47.22$\pm$1.79/-46.44$\pm$0.23 & -44.72$\pm$0.82/-43.38$\pm$0.63 & -42.80$\pm$1.85 & -35.02$\pm$0.49 \\ 
 & \textbf{DGRU} & \textbf{-41.85$\pm$0.51} & \textbf{-49.20$\pm$0.64/-46.44$\pm$0.67} & \textbf{-44.69$\pm$0.57/-44.47$\pm$0.27} & \textbf{-44.20$\pm$0.56} & \textbf{-35.22$\pm$0.38}\\ \hline
\end{tabular}
\begin{tablenotes}
\item[a] The numbers of parameters are 495 (GMP), 488 (GRU), 488 (LSTM), 538 (VDLSTM), 500 (RVTDCNN), and 486 (DGRU).
\item[b] The best PA modeling NMSE values over 100 epochs on the test set are -20.85\,dB (GMP), -30.85\,dB (GRU), -31.02\,dB (LSTM), -28.79\,dB (VDLSTM), -28.76\,dB (RVTDCNN), and -31.61\,dB (DGRU). DPD models in this table were all trained in pairs with the DGRU PA model to ensure the same initial condition in DPD learning.
\item[c] Calculated from the DGRU PA models with 5 different random seeds.
\item[d] The extracted features include \(I_x\), \(Q_x\), \(|\mathbf{x}|\), \(|\mathbf{x}|^3\), \(\sin\theta_x\), \(\cos\theta_x\).
\end{tablenotes}
\end{threeparttable}
\end{table*}

The collected dataset named \texttt{DPA\_200MHz} comprises 38,400 samples of 200\,MHz OFDM signals sampled at an 800\,MHz rate. The training is done using the \texttt{ADAMW}~\cite{kingma2014adam,loshchilov2018decoupled} optimizer for 100 epochs with the \texttt{ReduceLROnPlateau} scheduler, starting at $1\times 10^{-3}$ with a batch size of 64, a frame length of 50 and stride of 1. 

The performance of our proposed DGRU model in PA modeling and DPD learning was benchmarked against baseline models GMP, GRU, and LSTM, complemented by recent works VDLSTM~\cite{VDLSTM} and RVTDCNN~\cite{Hu2022RVTDCNN}, chosen for their architectures that entail relatively less domain-specific presumptions, making them apt for E2E learning. Except for the DPD parameter scan experiments, all PA and DPD models have around 500 parameters, and only the DPD model is needed during deployment. The input features of baseline GRU and LSTM models are $I_\mathbf{x}$, $Q_\mathbf{x}$, while VDLSTM and RVTDCNN used their best configuration reported in \cite{VDLSTM} and \cite{Hu2022RVTDCNN}, where their model sizes were respectively adjusted by increasing the hidden LSTM layer and the hidden FC layer. 
We trained and measured all models five times with distinct random seeds to account for training stochasticity, reporting mean metrics and corresponding standard deviation values.

\subsection{Results and Discussion}
\begin{figure}[t]
    \centering
    \includegraphics[width=1.0\linewidth]{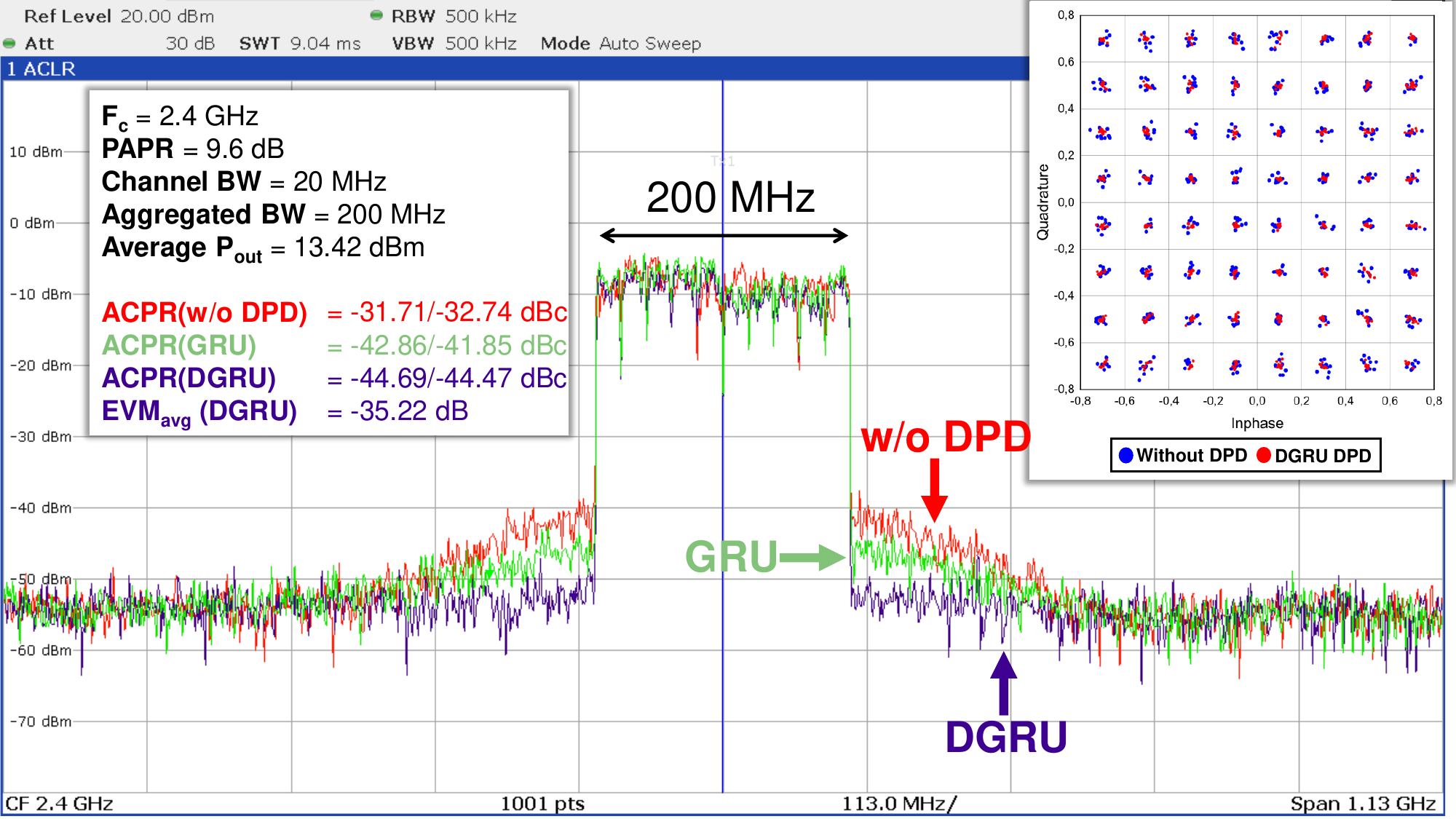}
    \caption{Measured spectrum and constellation map on the 200\,MHz Signal.}
    \label{fig:PSD}
\end{figure}
As illustrated in Fig.~\ref{fig:ACPR}(a), the NMSE reduces with training iterations during the PA modeling process. DGRU demonstrates superior accuracy among the various algorithms by yielding the minimum error on the validation set data after 100 epochs. To standardize the initial conditions for DPD learning across all algorithms, the gold DGRU PA model with the lowest NMSE over 100 epochs was adopted.

Figs.~\ref{fig:ACPR}(b) and (d) delineate the SIM-ACPR and SIM-EVM on the validation set across 100 epochs, respectively. The selection of the gold DPD model per algorithm was based on the lowest mean SIM-ACPR throughout the epochs. The neural network-based models notably outperformed the GMP in linearizing the wideband DPA. Furthermore, the DGRU attained the best SIM-ACPR and SIM-EVM after training.

Figs.~\ref{fig:ACPR}(c) and (e) further reveal the measured ACPR and EVM across a parameter range spanning from 100 to 3200 measured on the gold DPD model of each algorithm. The DGRU models show better performance with a mere excess of 128 parameters. As the parameter count escalates, the DGRU-DPD consistently outperforms other algorithms.

Moreover, the comparison between the SIM-ACPR, SIM-EVM curves, and their measured counterparts with around 500 parameters (vertical dashed lines in Figs.~\ref{fig:ACPR}(c) and (e)) exhibits a reliable congruence regarding Measured-ACPR and Measured-EVM performance rankings. For instance, simulated results accurately reflect the measurement-derived left (L) and right (R) ACPR for each DPD model. They also capture the performance hierarchy among different algorithms, indicating that the simulated metrics could be a trustworthy proxy for evaluating DPD performance in real-world applications.

Table~\ref{tab:200} summarizes the performance of DPD models with around 500 parameters on the $\texttt{DPA\_200MHz}$ test set. The findings underscore the effectiveness of the E2E learning architecture in training DNN-based DPD models such as VDLSTM, RVTDCNN, and even baseline GRU and LSTM models to accomplish commendable linearization with ACPRs surpassing -41\,dBc and EVMs around -34\,dB. The DGRU-DPD is the top performer, attaining an ACPR of -44.69/-44.47\,dBc, and an EVM of -35.22\,dB.

Table~\ref{tab:200} also delineates the incremental performance gains achieved by DGRU by integrating an FEx layer and implementing a dense skip path. By replacing the basic $I_\mathbf{x}$ and $Q_\mathbf{x}$ inputs with their feature-extracted counterparts while retaining the GRU backbone, an improvement in ACPR, from -42.86/-41.85\,dBc to -44.72/-43.38\,dBc, was observed. The transition to the DGRU backbone leads to the best values in both simulated and measured ACPR and EVM.

Finally, Fig.~\ref{fig:PSD} shows the measured spectrum and the constellation diagrams for PA outputs, comparing scenarios with and without DPD, underscoring the capability of \texttt{OpenDPD} in training DNN-based DPD models to correct non-linearity in a wideband DPA effectively.
\section{Conclusion}
\label{sec:conclusion}
This paper introduces the \texttt{OpenDPD} framework, aiming to streamline DPD exploration and standardize datasets for fair comparisons. Results show our E2E learning architecture can effectively train various DNN-based models for DPD purposes. Our proposed DGRU achieved superior results in the benchmark, exceeding both baselines and prior models. We plan to regularly update \texttt{OpenDPD} with new analog and digital PA datasets and pre-trained models with the latest DNN backbones and welcome interdisciplinary contributions.

\bibliographystyle{IEEEtran}
\bibliography{ref.bib}

\end{document}